\documentclass[runningheads]{llncs}

\usepackage{graphicx}
\usepackage{xcolor}
\usepackage{marvosym}

\begin{document}

\title{A Smartphone-based System for Real-time\\Early Childhood Caries Diagnosis}

\titlerunning{A Smartphone-based System for Real-time ECC Diagnosis}

\author{Yipeng Zhang\inst{1}\textsuperscript{(\Letter)} 
\and Haofu Liao\inst{1} 
\and Jin Xiao\inst{2} 
\and Nisreen Al Jallad\inst{2}
\and Oriana Ly-Mapes\inst{2}
\and Jiebo Luo\inst{1}}

\index{Zhang, Yipeng}
\index{Liao, Haofu}
\index{Xiao, Jin}
\index{Al Jallad, Nisreen}
\index{Ly-Mapes, Oriana}
\index{Luo, Jiebo}

\authorrunning{Zhang et al.}
\institute{Department of Computer Science, University of Rochester \and 
Eastman Institute for Oral Health, University of Rochester Medical Center
\\
\email{yzh232@u.rochester.edu}}

\maketitle              

\begin{abstract}
Early childhood caries (ECC) is the most common, yet preventable chronic disease in children under the age of 6. Treatments on severe ECC are extremely expensive and unaffordable for socioeconomically disadvantaged families. The identification of ECC in an early stage usually requires expertise in the field, and hence is often ignored by parents. Therefore, early prevention strategies and easy-to-adopt diagnosis techniques are desired. In this study, we propose a multistage deep learning-based system for cavity detection. We create a dataset containing RGB oral images labeled manually by dental practitioners. We then investigate the effectiveness of different deep learning models on the dataset. Furthermore, we integrate the deep learning system into an easy-to-use mobile application that can diagnose ECC from an early stage and provide real-time results to untrained users. 

\end{abstract}

\section{Introduction}
Early childhood caries (ECC) is defined as the presence of $\geq1$ decayed, missing (due to caries), or filled tooth surface in primary teeth in a child 71 months of age or younger \cite{ccolak2013early}. ECC is by far the most common chronic childhood disease, with nearly 1.8 billion new cases per year globally \cite{dye2007trends,dye2012oral,vos2017global}. In the US, it afflicts approximately 28\% of children aged 2-5 years while disproportionately affecting up to 54\% of low-income and minority children \cite{dye2007trends,office2000us}. If not treated properly in time, the decay may become more severe, where usually a total oral rehabilitation treatment (TOR) under general anesthesia \cite{koo2014candida} with multiple tooth extractions and restorations/crowns is required, at a cost of nearly \$7,000 per child (2009-2011 data) \cite{rashewsky2012time}, leading to a USD 1.5 billion expense on the treatment per year in USA \cite{xiao2019prenatal}. Further, almost 40\% of children experience recurrent caries 6-month post TOR \cite{berkowitz2011dental,graves2004clinical}. Untreated ECC often leads to higher risk of caries lesions in permanent teeth, diminished oral health-related quality of life, hospitalizations and emergency room visits due to systemic infection, and even death \cite{american2009american,casamassimo2009beyond}. 

Continued use of these expensive restorative methods is a drain on the health care system. Despite years of clinical and research endeavors, the challenge remains, primarily due to the fact that, without training, it is hard for the general population to identify the disease from its appearance at its early stages. The current biomedical approach to control the ECC pandemic primarily focuses on individual-level restorative procedures rather than population-wide preventive strategies. Many US preschool children from low-income family often have poor access to pediatric dental services, leaving them in a situation that dental caries often are not diagnosed until at the late stages. Moreover, ECC is a multifactorial disease with host, microorganisms, diet and oral hygiene practice as the factors that determine the risks, children’s caregivers need to be educated extensively about these risk factors that can be self-managed and altered in order to reduce children’s risk for ECC. 

In recent years, deep convolutional neural networks (CNN) have demonstrated supra-human level performance in many classification and object detection tasks \cite{he2015delving,zhao2019object}. Previous studies have also shown the efficacy of CNNs on the task of classifying dental caries on per-tooth periapical radiographs \cite{lee2018detection}. However, to the best of our knowledge, no deep learning-based method has been proposed for detecting dental caries from RGB images captured by regular cameras, as well as any  suitable dataset for this task. In this study, we adopt an object detection approach. There are currently many one-stage object detection algorithms that are capable of performing fast, real-time inference while maintaining comparable accuracy \cite{liu2016ssd,redmon2018yolov3} in company with light-weight feature extractor networks \cite{ma2018shufflenet,sandler2018mobilenetv2}. Multistage object detectors \cite{ren2015faster} with larger feature extractor networks \cite{he2016deep,simonyan2014very} often yield better accuracy but require longer inference time. In this study, we compare the performance of selected frameworks on the cavity detection task.

The training of fully supervised object detection models requires a large amount of fully labeled image data. However, the collection of medical data is challenging considering the number of cases and sensitive information. Nevertheless, we collect 1,000 oral images to aid the training. We classify each tooth into 8 distinct categories and create a reasonably large-scale intraoral image dataset that includes the oral images with a total of 15,692 bounding boxes. All images in the dataset are annotated manually by dental practitioners. To account for the multifactorial nature of ECC, we propose a multistage deep learning-based diagnostic system that utilizes both analysis of user habits and symptoms and detection results from a cavity detector trained on our dataset. Further, we deploy the deep learning system onto Android devices to provide the highest ease of use. The deployed system provides caregivers, especially among the underserved population, a first-hand tool to detect early stages of ECC. The system also contains an interactive education module to provide caregivers extensive knowledge on ECC prevention. Experimental results demonstrate that the system proposed has the capability and potential to break the current ECC prevention deadlock.

In summary, our main contributions are four folds: 
\begin{itemize}
\vspace{-2mm}
    \item We propose a multistage deep learning-based diagnostic method that employs both analysis of user habits and symptoms and detection results by a cavity detector trained on a large-scale expert-annotated intraoral image dataset. 
    \vspace{2mm}
    \item We integrate image capture assistance and automated image analysis into a streamlined process on Android devices to provide the highest ease of use to family caregivers who are untrained and not technology-savvy. 
       \vspace{2mm}
    \item We incorporate an interactive education module to provide family caregivers extensive knowledge on ECC prevention. 
       \vspace{2mm}
    \item We obtain experimental results to demonstrate that the entire smartphone-based system has the capability and potential to break the bottleneck in current ECC prevention.
\end{itemize}

\section{Method}

\begin{figure}
\includegraphics[width=1\textwidth]{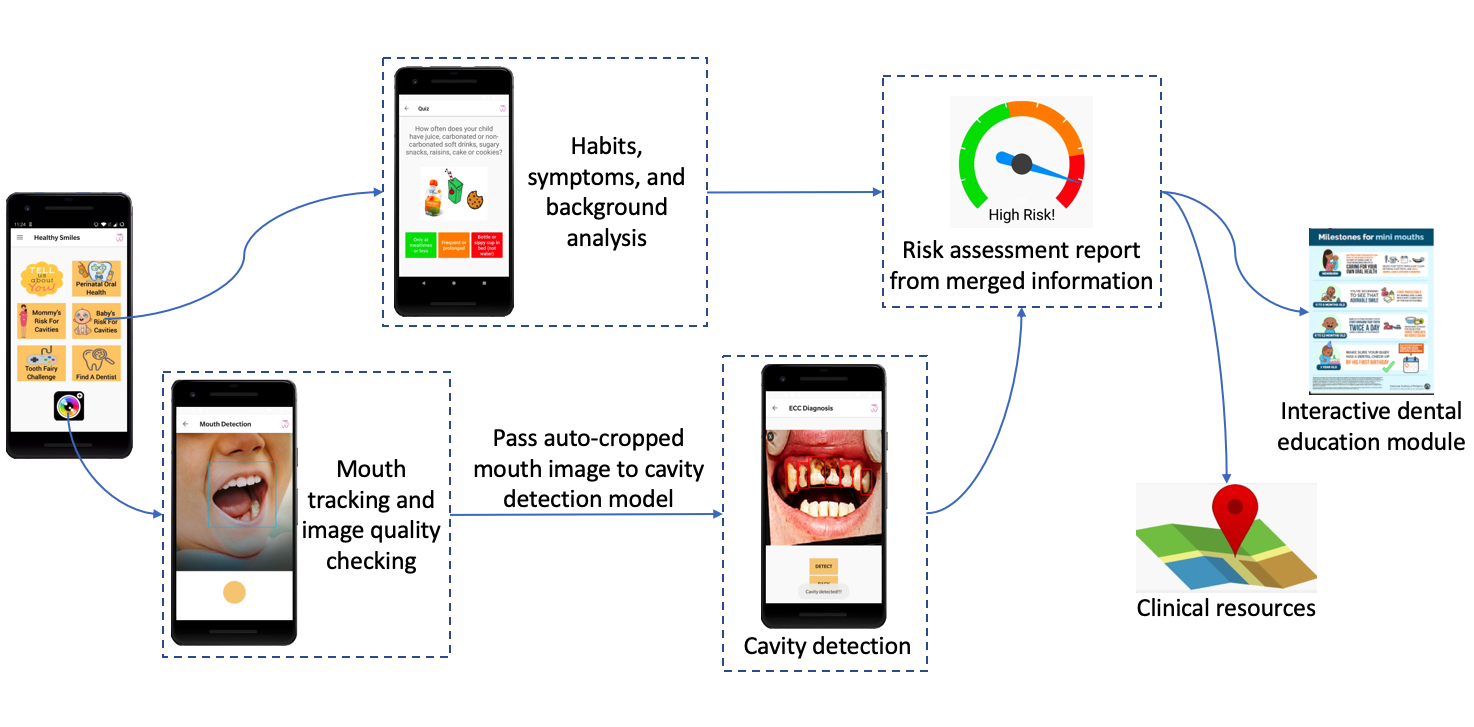}
\caption{Overview of the smartphone-based caries diagnosis system. Note that the interface pictures are actual screenshots from Android devices. (icon source: \cite{aap,pixabay})} \label{fig1}
\end{figure}

The proposed diagnostic system consists of two main modules: a 2-stage visual cavity detection module and a multifactorial analyzer module as shown in Fig. 1. The cavity detection module detects the visual signs of dental caries from a user-taken photo by a standard smartphone. The multifactorial analyzer module gathers information about the various factors that might affect the formation of caries, including the patient's habits, diet, and symptoms through interactive question answering and calculates the potential risk. Then, the risk assessment module takes into account results from both modules and generates an informative report. From there the user can choose to view the nearby certificated dentists' information and other clinical resources, or to enter an interactive education module that incorporates dental knowledge into a textual mini-game that aims to increase the family caregivers' knowledge about early signs and preventative methods of ECC.

\subsubsection{Mouth Tracker}
We deploy a real-time facial landmark detector to perform mouth tracking when the user is in the camera page. Based on the landmark points, our model visualizes a mouth bounding box and calculates the orientation of the mouth and quality of the image. The device will notify the user if the bounded mouth is not satisfactory for the following cavity detection (e.g. image quality is too low ($<224 \times 224$), mouth is tilted too much ($>5^\circ$), etc.). Otherwise, the user is prompted to take the photo. The mouth detected is then cropped and passed to the cavity detector.

\subsubsection{Cavity Detector}
We choose the Single Shot Multibox Detector (SSD) \cite{liu2016ssd} using a  MobileNetV2 backbone \cite{sandler2018mobilenetv2} as our cavity detection model. MobileNetV2 is an efficient CNN architecture that utilizes depth-wise separable convolutions to increase the speed of inference. This architecture is suitable for our task of performing real-time inference on smartphones, since those devices usually do not have very powerful computing hardware  (e.g. GPUs) that we can utilize to accelerate tensor operations. We also compare its performance on our dataset with the much larger Faster R-CNN \cite{ren2015faster} using a ResNet50 backbone \cite{he2016deep}, and with trained dentists.

\subsubsection{Risk Assessment}
The risk assessment module takes into account primarily visual cavity detection results. If severe cavity (\textit{level2}, see the Experiment section for detailed classification rules) is detected, the user is prompted to seek guidance from dental professionals immediately. If no cavity or early signs of cavity (\textit{level1}) is detected, we take the result from the multifactorial analyzer module and inform the user as such. The result is based on the the American Dental Association Caries Risk Assessment system \cite{young2015american}. We modify the elements in the system to accommodate the less health-literate individuals.

\subsubsection{Education and Clinical Resources}
We assembled a series of informative educational materials that provide appropriately timed information specific to pregnant women’s oral health importance (as they may affect oral healthy conditions of the to-be-born child), children’s tooth development, children’s oral hygiene and diet recommendations. The educational information is built into forms of interactive mini-games like scored question answering, literacy bubble shooter, and son. We also provide a list of available dental clinics near the current location that accept dental insurance for low-income group (e.g. Medicaid).

\subsubsection{Mobile Deployment}
Our goal is to help the low-income families that have limited access to  regular dental check-up. Therefore, we build the proposed deep learning system into an Android application in order to maximize both the coverage and accessibility. We use the light-weight SSD+MobileNetV2 model in the application for best user experience in real-time. Currently, the application functionality is successfully tested on Google Pixel 2 and OnePlus 5T with Android version $\geq$ 8.1. The deep learning models (mouth and cavity detector) perform inference at the speed of on the average 10 FPS on the tested devices.

\section{Experiment}
\subsection{Dataset}

\begin{figure}
\centering
\includegraphics[width=0.9\textwidth]{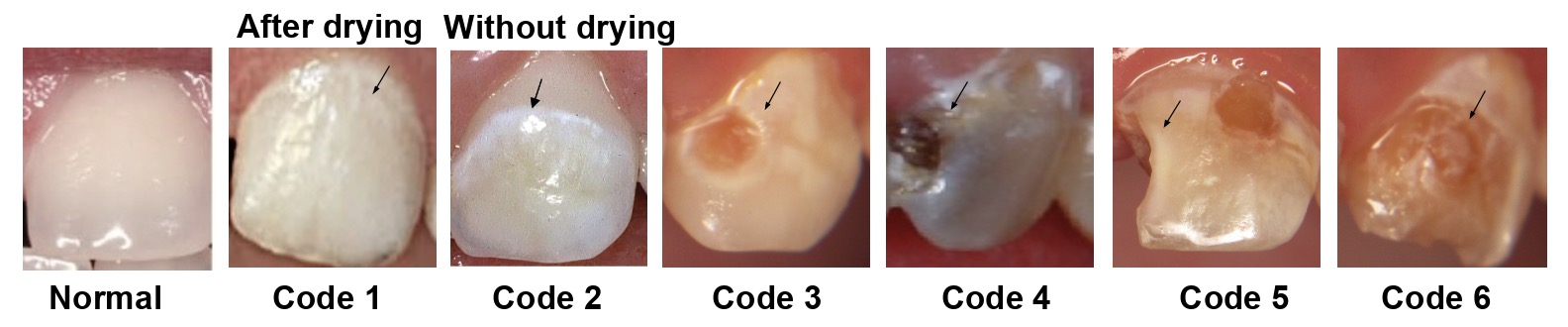}
\caption{Example images that demonstrate the ICDAS  visual diagnostic criteria. Tooth hard surface is composed of the outer enamel and inner dentin layers. Codes 1 and 2 indicate the initiation of white spots on non-cavitated tooth enamel indicated by the arrows, with or without air-drying. Codes 3 and 4 indicate caries on cavitated tooth enamel. Code 4 reveals the gray shadow around and underneath the enamel. Codes 5 and 6 indicate caries involving dentin, with ascending extensiveness from 5 to 6.} \label{fig2}
\end{figure}

We adopt the International Caries Diagnosis System (ICDAS) \cite{gugnani2011international} as the reference on the severity of cavity. ICDAS is the gold standard caries diagnostic index for scoring caries severity and routinely used in caries research and epidemiological studies \cite{shoaib2009validity}. Following the criteria, we create the intraoral image dataset that consists of 1,000 fully labeled high resolution RGB oral images from consented patients, each containing around 15 tooth bounding boxes. 

Our current work focuses on the anterior teeth since the anterior teeth erupt earlier than the posterior teeth and the maxillary anterior teeth are the most affected by ECC. Moreover, it is less technically challenging for parents to take front teeth photos using smartphones. Of the teeth photos in our dataset, 90\% were obtained from children younger than 18 years of age, including 80\% permanent anterior teeth and 10\% primary anterior teeth. The enamel texture and shape of the permanent and primary teeth are similar, which do not introduce bias in caries detection and categorization using the ICDAS standard, whose categories are based on lesion color and size.

Besides the ICDAS categories, we add an \textit{other} label to include human-imposed devices that block the teeth surface (e.g. braces, steel crowns). In total, there are 15,692 bounding boxes from 8 classes (numbers in the parentheses denote the number of each type in the dataset): \textit{normal}(6,825), \textit{code1}(0), \textit{code2}(6,832), \textit{code3}(990), \textit{code4}(33), \textit{code5}(382), \textit{code6}(189), \textit{other}(441). There are no tooth labeled as code 1 because code 1 requires drying of the tooth surface, which is not indicated by the image. Due to the similarity of codes 1 and 2, we labeled all the teeth with code 1 or 2 symptoms as code 2. The usage of this dataset can be generalized to general cavity diagnosis studies since the symptoms of childhood caries and caries at other ages are very similar.

\begin{figure}
\centering
\includegraphics[width=0.8\textwidth]{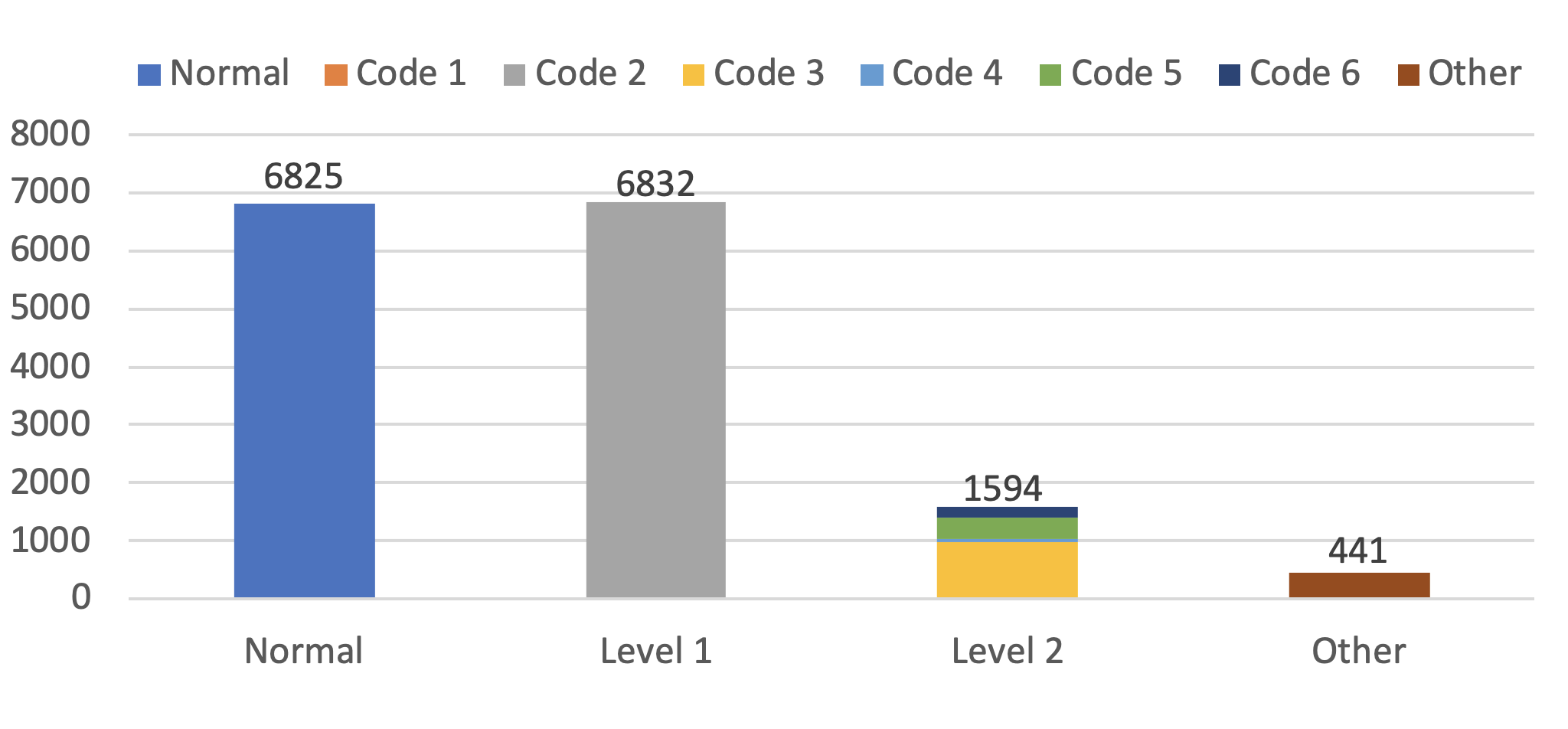}
\caption{Class distribution and the grouping schema of the intraoral dataset.} \label{fig3}
\end{figure}

As shown in Fig. 2, most people are able to identify the cavity once the teeth surface starts to decay at code 3 and onward. However, it is challenging for the untrained general population to spot the cavity at codes 1 and 2. Therefore, in our task, it is sufficient to group codes 1 and 2 as category \textit{level1} that consists of teeth at early stages of cavity, and codes 3-6 as category \textit{level2} that consists of heavily cavitated teeth. The remaining 2 categories (\textit{normal} and \textit{other}) remain the same. The resulting distribution is shown in Fig. 3. We use the new 4-category dataset to train our object detection models and apply class weighting to deal with the imbalance.

\subsection{Evaluation}

We resize the images in our dataset to size 224x224 and split the dataset into training-validation set (80\%) and test set (20\%). We train both SSD and Faster R-CNN models using the SGD optimizer with a learning rate of $2e^{-4}$ and batch size of 8. The models are trained for 150 epochs of the training data. We evaluate the performance of the 2 models qualitatively and quantitatively. We also evaluate the performance of the deployed SSD-MobileNetV2 model with trained dentists.

\begin{figure}
\includegraphics[width=1\textwidth]{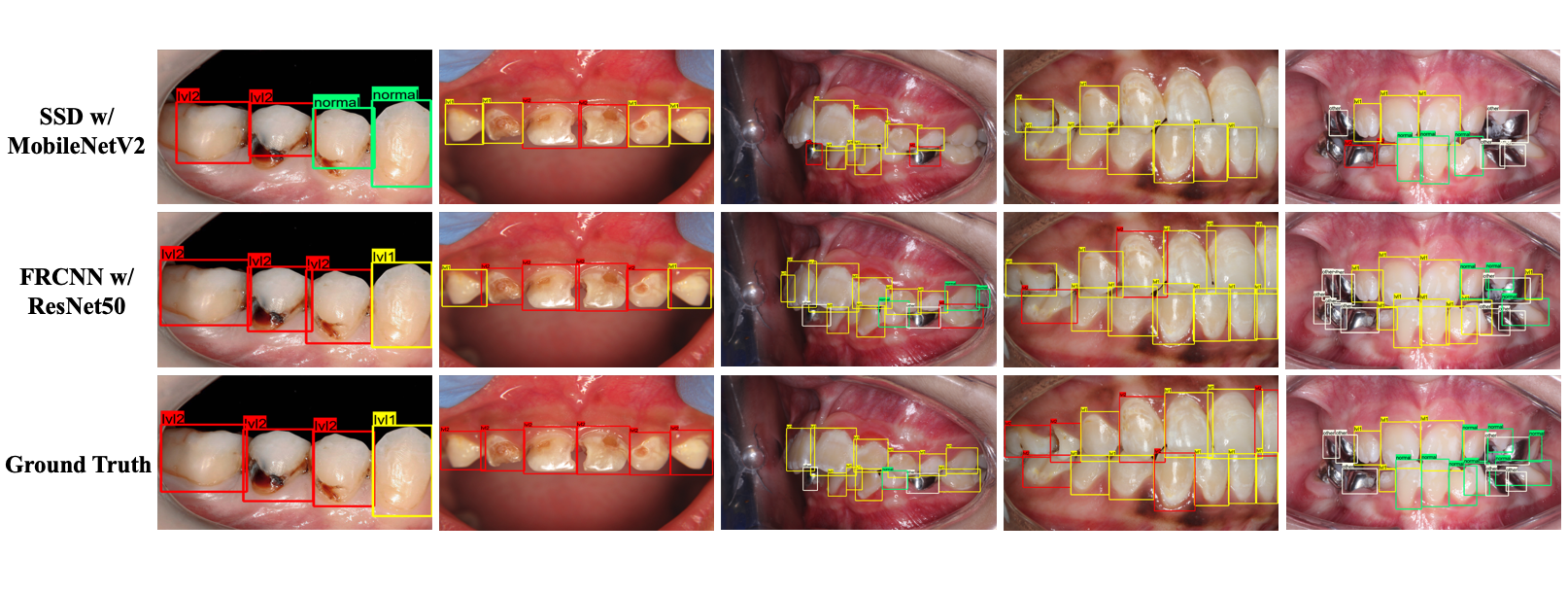}
\caption{Qualitative comparison between the cavity detection results by SSD and Faster R-CNN. Green boxes denote normal teeth; yellow boxes denote level 1 cavity; red boxes denote level 2 cavity; and white boxes denote human-imposed oral devices.} \label{fig4}
\end{figure}

\subsubsection{Qualitative Comparison}
Fig. 4 shows the representative qualitative detection result comparison on the test set. We choose the images containing a wide range of teeth categories to visualize the models' performance on all categories. It is clear that both models are able to correctly detect and classify most of the teeth in each image. We find the Faster R-CNN model performs significantly better than the SSD model; it successfully locates almost all the teeth and the misclassified teeth are sometimes those that are ambiguous in terms of severity levels. This is possible due to the different lighting conditions of the images. For example, the white spots in the signal \textit{level1} cavity class can be caused by flashlight from the camera. We observe that the SSD model sometimes cannot detect all the teeth in an image when the number of teeth is large. However, it also shows the capability of detecting most teeth that contain caries.

\begin{table}
\setlength{\tabcolsep}{8pt}
\renewcommand{\arraystretch}{1.2}
\centering
\caption{Quantitative detection results on our dataset (MSCOCO metrics).}\label{tab1}
\begin{tabular}{ c | c | c | c | c | c} 
\hline
 &  $mAP$ & $AP_{50}$ & $AP_{75}$ & $AR$ & Speed (ms) \\
\hline
\bfseries SSD-MobileNetV2 &  0.303 & 0.522 & 0.320 & 0.562 & 22.45\\
\bfseries Faster-RCNN-ResNet50 &  0.473 & 0.665 & 0.543 & 0.716 & 91.66\\
\hline
\end{tabular}
\end{table}

\subsubsection{Quantitative Comparison}
Table 1 shows the performance scores of the two models using the MSCOCO metrics \cite{lin2014microsoft}. We also measure the average inference time per image on a Tesla K80 GPU. We present 4 major score fields: mean average precision ($mAP$), average precision at 0.5 and 0.75 intersection over union(IoU) thresholds ($AP_{50}$, $AP_{75}$ respectively), and average recall ($AR$). We focus mostly on $AR$ and $AP_{50}$. Since our goal is to spot the cavity and prompt the user to seek help from dentists, a high IoU score is much less important compared to ability of detecting the disease. Faster R-CNN still outperforms the SSD model in all the score fields, while taking much longer for inference.

\begin{figure}
\begin{centering}
\includegraphics[width=0.6\textwidth]{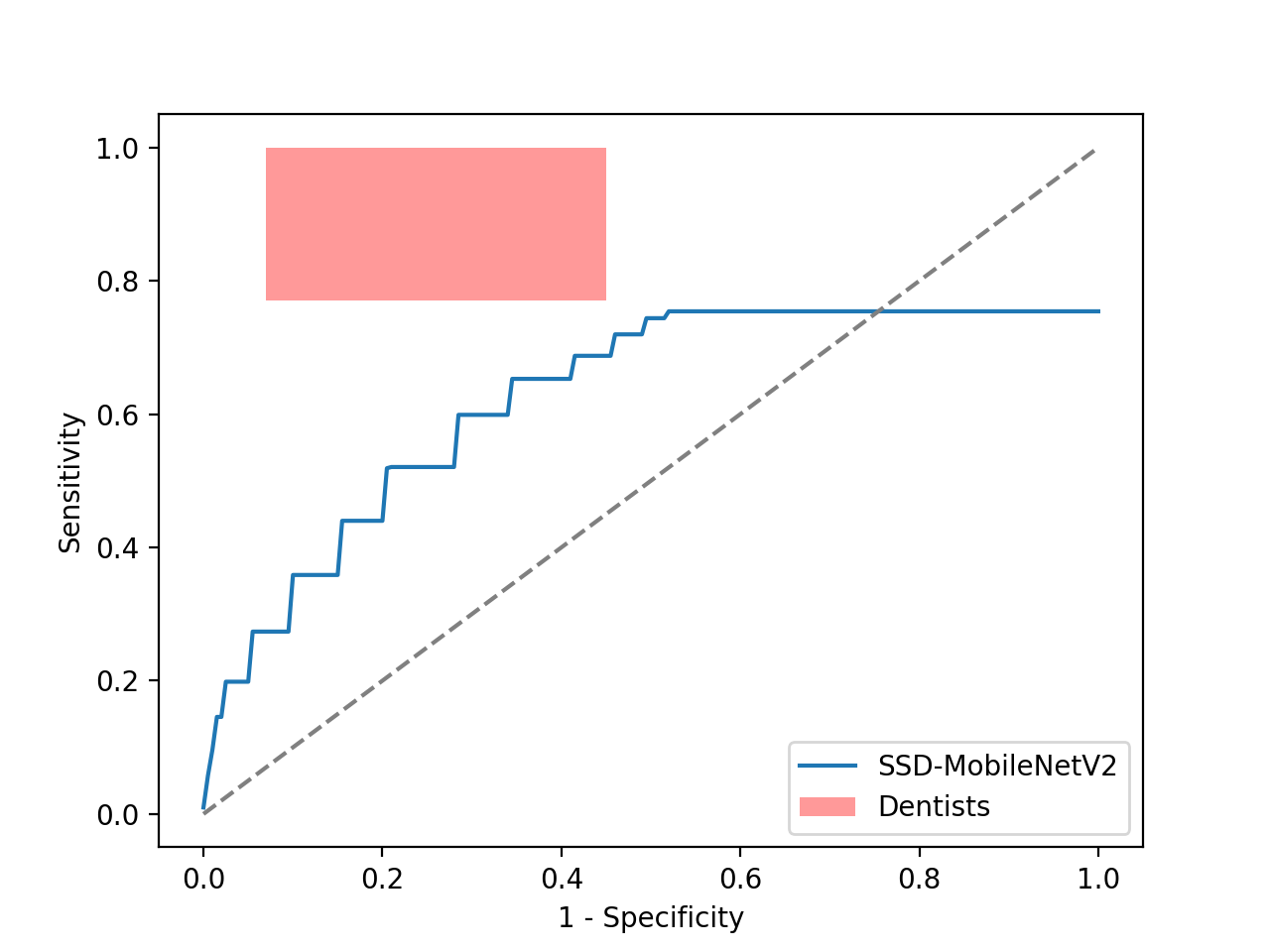}
\caption{ROC curve of cavity detection by the SSD-MobileNetV2 model. The red box denotes the average performance of dentists. The curve does not reach the upper right corner because we take into account the undetected teeth and incorrect detections.} \label{fig5}
\end{centering}
\end{figure}

\subsubsection{Model-Dentist Comparison}
Currently, we incorporate the SSD model into the mobile application since it is fast and lightweight, thus not requiring high-end computational-expensive devices that our targeted audiences cannot afford. We then evaluate the performance of our system on medical metrics using the ROC curve. We evaluate the model by deriving an $(K + 1) \times (K + 1)$ detection confusion matrix ($K$ denotes number of classes, 4 in our case) from the detection results on the test set. The extra row and column take into account the undetected teeth and incorrect detections (detected unlabeled areas). We then plot the ROC curve in Fig.\ref{fig5} by calculating from the confusion matrix the sensitivity and specificity scores with respect to cavity detection instead of per category detection. We achieve this by grouping \textit{level1} and \textit{level2} as one category namely \textit{cavity} and the rest as another. Previous studies have estimated the average performance of dentists on identification of dental caries: 0.77 to 1.00 for sensitivity and 0.45 to 0.93 for specificity \cite{gordan2011methods}. The performance of dentists is visualized on the plot. We notice that there is still some distance between the performance of our deployed SSD model and that of trained dentists, but the gap is expected to shrink if we deploy a larger model or label more data for training. 

We believe that the current system already offers a good addition to the current dental healthcare practice by allowing untrained family caregivers to participate, with the help of computerize image recognition, in the effort to break the bottleneck in ECC diagnosis and prevention. With the status quo, many ECC cases are simply undetected until it is too late and too costly to treat.

\section{Conclusion and Future Work}

We present a smartphone-based system for real-time early childhood caries diagnosis to fill in a major gap in ECC diagnosis and prevention in the current health care system. We create a first-ever large-scale intraoral dataset labeled manually by dental practitioners. The dataset can be used for future research not only on ECC but also on the general cavity diagnosis. We design a multistage deep learning system that uses object detection models supported by multifactorial user information analysis to perform both risk assessment and dental education. Furthermore, we deploy the system on Android devices to test its functionality and feasibility. In the future, we will perform clinical testing to evaluate the impact of the deployed system, while investigating ways to further improve the performance and deployability of other deep learning models. We will also extend the detection model to posterior teeth diagnosis using our archived images.

\bibliographystyle{splncs04}
\bibliography{mybibliography}

\end{document}